\documentclass[sigplan,screen,nonacm]{acmart}

\AtBeginDocument{%
  \providecommand\BibTeX{{%
    \normalfont B\kern-0.5em{\scshape i\kern-0.25em b}\kern-0.8em\TeX}}}

\usepackage{amsmath} 
\begin{document}
\title{Toward Fairness via Maximum Mean Discrepancy Regularization on Logits Space}


\author{Hao-Wei Chung}
\authornote{Both authors contributed equally to this research.}
\email{xdmanwww@gapp.nthu.edu.tw}
\affiliation{%
  \institution{National Tsing Hua Unviserity}
  \city{Hsinchiu}
  \country{Taiwan}
}

\author{Ching-Hao Chiu}
\authornotemark[1]
\email{gwjh101708@gapp.nthu.edu.tw}
\affiliation{%
  \institution{National Tsing Hua Unviserity}
  \city{Hsinchiu}
  \country{Taiwan}
}

\author{Yu-Jen Chen}
\affiliation{%
  \institution{National Tsing Hua Unviserity}
  \city{Hsinchiu}
  \country{Taiwan}
\email{yujenchen@gapp.nthu.edu.tw}
}

\author{Yiyu Shi}
\affiliation{%
  \institution{University of Notre Dame}
  \city{Notre Dame}
  \state{Indiana}
  \country{USA}
\email{yshi4@nd.edu}
}

\author{Tsung-Yi Ho}
\affiliation{%
  \institution{National Tsing Hua Unviserity}
  \city{Hsinchiu}
  \country{Taiwan}
\email{tyho@cs.nthu.edu.tw}
}


\begin{abstract}
Fairness has become increasingly pivotal in machine learning for high-risk applications such as machine learning in healthcare and facial recognition. However, we see the deficiency in the previous logits space constraint methods. Therefore, we propose a novel framework, \textbf{Logits-MMD}, that achieves the fairness condition by imposing constraints on output logits with Maximum Mean Discrepancy. Moreover, quantitative analysis and experimental results show that our framework has a better property that outperforms previous methods and achieves state-of-the-art on two facial recognition datasets and one animal dataset. Finally, we show experimental results and demonstrate that our debias approach achieves the fairness condition effectively.
\end{abstract}






\maketitle
\section{Introduction}
\label{sec:intro}
\begin{figure}[h]
\begin{center}
\includegraphics[width=1.0\linewidth]{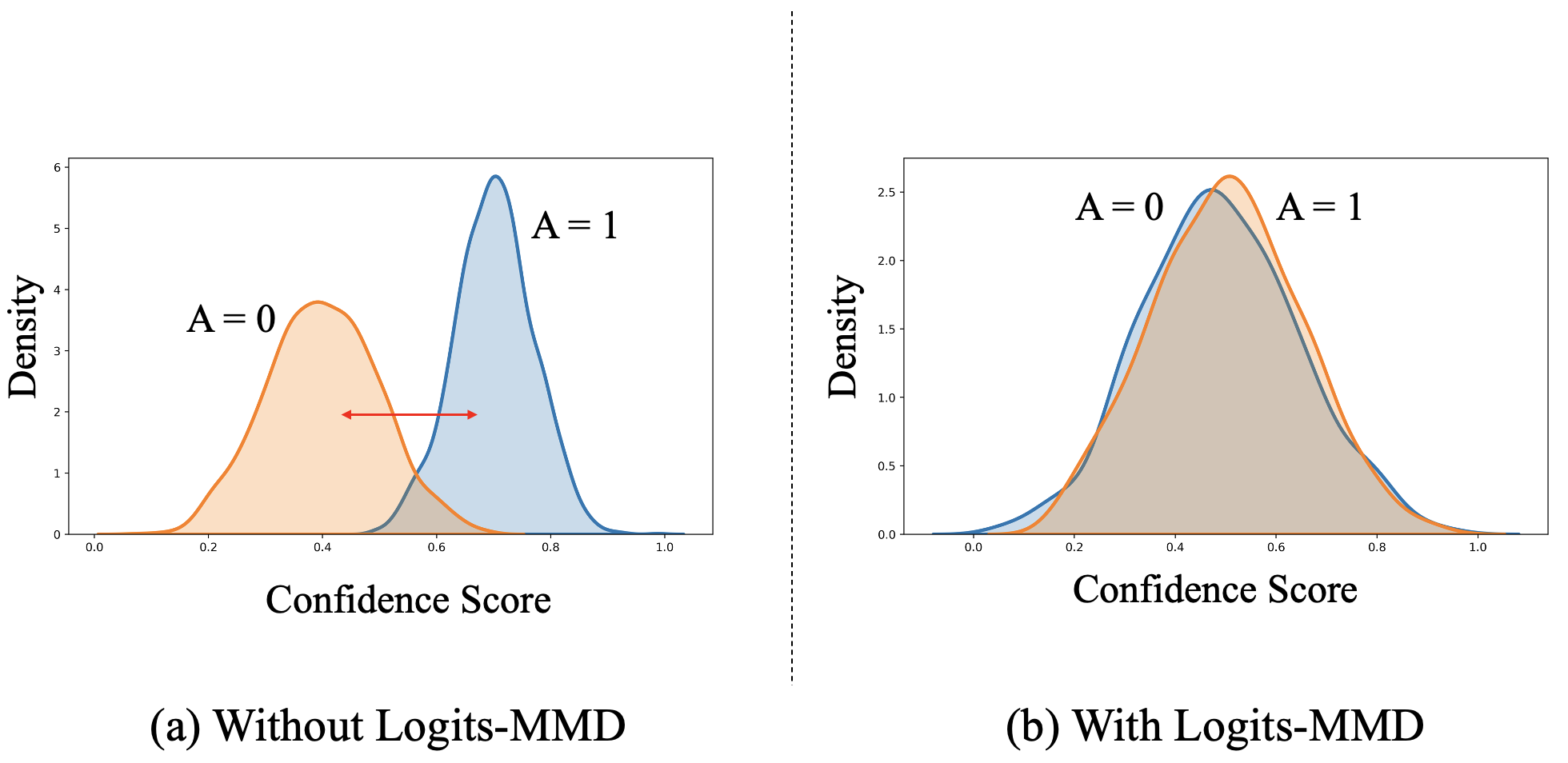}
\caption{The Logits-MMD constraint pulls the logits distribution of different sensitive groups together i.e. \emph{A} = 0 and \emph{A} = 1. (a) is an example of logits distribution without Logits-MMD constraint. (b) is an example of imposing Logits-MMD constraint.}
\label{Fig.intro}
\end{center}
\end{figure}

Due to the power of the neural network, computer vision was endowed with the capability to impact society significantly. As a result, many high-risk computer vision applications emerged, such as medical diagnosis and facial recognition. Due to these applications, it is pivotal to make sure the deployed model makes a ``fair'' prediction; specifically, the prediction should be independent of the sensitive attributes of the users, such as race, gender, and age. However, automated systems that use visual information to make decisions are vulnerable to data bias. 


Several methods are proposed to alleviate the bias in machine learning models. Many of them \cite{wang2022fairness, kim2019learning} adopted adversarial training to eliminate bias by training the network to learn a classifier while disabling the adversary's ability to categorize the sensitive attribute. There was also a mainstream that used regularization terms \cite{romano2020achieving} in the objective to enforce the model learning the fair representation. In this work, we propose a debias method that imposes constraints on the classifier during the training phase to achieve fairness.


Most of the debias work evaluated the fairness condition by the prevalent metric equalized odds (EO) \cite{hardt2016equality}, which measures the difference between the true positive rate (TPR) and the false positive rate (FPR) against different sensitive groups. A lower EO indicates the greater fairness of a model. Based on the definition, we can reformulate the EO equation by replacing the TPR and FPR with the integral of the model confidence score's probability density function, where confidence score is the value of the logits after applying the sigmoid function. Illustrated in Fig. \ref{Fig.intro}, we propose a novel fairness method \textbf{Logits-MMD} to train a fair classifier that takes EO under consideration while training. With the above intuition, \textbf{Logits-MMD} minimizes EO by minimizing the distribution distance of the classifier's output logits in terms of TPR and FPR between different sensitive groups via Maximum Mean Discrepancy (MMD) \cite{gretton2012kernel}. Moreover, we point out the drawback of the similar logits space regularization methods,``Gaussian Assumption'' and ``Histogram Approximation'' proposed by \cite{chen2020towards}, and demonstrate that the objective of the previous two methods are not consistent with the fairness objective while our MMD based regularizer has a better property of minimizing EO. We conduct facial attribute classification on the CelebA \cite{liu2018large} and UTK Face \cite{zhang2017age} datasets to validate our method. Our experiments demonstrate a 40.6\% improvement on average in terms of fairness compared to the state-of-the-art method in the CelebA dataset and a 13\% improvement on average in the UTK Face dataset. Moreover, to demonstrate the effectiveness of our fairness methods on the more general bias level, we perform classification on the Dogs and Cats dataset \cite{dogNcat} where color is the sensitive attribute followed by \cite{kim2019learning} in the supplementary materials.

In the rest of the paper, Section 2 reviews the related work and outlines our motivation behind the research. Section 3 defines the problem and introduces \textbf{Logits-MMD}. Experimental settings and results are reported in Sections 4 and 5, while the conclusion is summarized in Section 6.



\section{Background and Motivation}
\label{sec:background}
\subsection{Related Work}
The bias mitigation methods are designed to reduce the native bias in the dataset to increase the chances of fair predictions. There are essentially three avenues for the current debiased strategy, including pre-processing, in-processing, and post-processing. 


\textbf{Pre-processing} strategies usually remove the information which may cause ``discrimination'' from training data before training. \cite{creager2019flexibly} achieved fairness by removing unfair information before training. \cite{lu2020gender} achieved fairness by pre-processing data, including data generation and augmentation.

\textbf{In-processing} methods consider fairness constraints during training. These studies modified existing training frameworks \cite{park2021learning, park2022fair, chiu2023toward, chiu2023fair} or incorporated a regularization term \cite{chen2020towards, jung2021fair, quadrianto2019discovering, romano2020achieving, chiu2024achieve} to achieve fairness goals. Adversarial training \cite{kim2019learning} is also a common technique that mitigates bias by adversarially training an encoder and classifier to learn a fair representation. Contrastive learning \cite{park2022fair} has also received extensive research in recent years. Recently, \cite{10.1145/3489517.3530427} proposed a fairness- and hardware-aware automatic neural architecture search (NAS) framework, the first work integrating fairness in NAS. Additionally, \cite{10247765} proposed a framework that simultaneously unites models to improve fairness on several attributes.  



\textbf{Post-processing} approaches aim to calibrate the model's output to enhance fairness. \cite{hardt2016equality} revealed the limit of demographic parity, and \cite{dwork2012fairness} gave a new metric to fairness equalized odds and showed how to adjust the learned prediction. Most recently, \cite{wang2022fairness} adversarially learned a perturb to mask out the input images' sensitive information.

\subsection{Fairness Metric}
\label{sec:fairness_def}
Consider a binary predictor \textit{p}, we apply the sigmoid function $\sigma$ to the output logits to obtain the confidence score $c \in [0, 1]$. Then, the model prediction $\hat{Y}$ will be obtained based on the given confidence threshold $t$.

\begin{equation}
\label{eq.EO}
\hat{Y}=\begin{cases}
1, & \text{ if } \sigma (p(x)) > t,\\
0, & \text{ else.}
\end{cases}
\end{equation}

Given the ground truth \textit{Y$\in$y} = $\left \{ 0, 1 \right \}$ and the predictor output \textit{$\hat{Y}$$\in$$\hat{y}$} = $\left \{ 0, 1 \right \}$, we have the fairness metrics for sensitive attribute \textit{$A$$\in$$a$} = $\left \{ 0, 1 \right \}$.

\textbf{Definition 1.} \textit{(Demographic Parity \cite{dwork2012fairness})} Demographic Parity (DP) consider the agreement of true positive rates between different sensitive groups. A smaller DP means a fairer model.
\begin{equation}
\label{eq.DP}
|P(\hat{Y}=1 \mid  A=0)-P(\hat{Y}=1 \mid A=1)|.
\end{equation}

\textbf{Definition 2.} \textit{(Equalized Odds \cite{hardt2016equality})} Equalized odds (EO) consider the true positive rates and false positive rates between different sensitive groups. A smaller EO means a fairer model.
\begin{equation}
\label{eq.EO_def_1}
\scalebox{0.95}[1]{$\sum_{\forall y \in \left \{ 0, 1 \right \}}\left|P(\hat{Y}=1 \mid A=0,Y=y)-P(\hat{Y}=1 \mid A=1,Y=y)\right|.$}
\end{equation}

While DP forces the same probability of being predicted as positive for each sensitive group, EO aims to achieve equal prediction accuracy across different sensitive groups so that both privileged and unprivileged groups have the same true positive rate and false positive rate, considering all classes. However, it's worth noting that DP is limited by the proportion of positive cases in different groups in the ground truth, making it less practical for real-world applications. As a result, EO is considered a more reasonable and flexible criterion.


Therefore, this paper focuses on achieving fairness by minimizing EO in multi-sensitive attribute settings with $A \in \{0,1,2,3,...,M\}$.

\textbf{Definition 3.} \textit{(Multi-sensitive attribute EO)} We extend EO (Eq. \ref{eq.EO_def_1}) to a multi-sensitive attributes manner following \cite{jung2022learning}.

\begin{equation}
\begin{gathered}
\label{eq.EO_def}
\sum_{\forall y \in \left \{ 0, 1 \right \}} \sum_{\forall a^i, a^j \in A}|P(\hat{Y}=1\! \mid\! A=a^i,Y=y)\!- \\ 
\!P(\hat{Y}=1\! \mid\! A=a^j,Y=y)|.
\end{gathered}
\end{equation}




\subsection{Equalized Odds Minimization}
\label{subsec:method_fairness_criterion}
To achieve fairness by minimizing EO (Eq. \ref{eq.EO_def}),  we first define $P_{a,y}(\hat{Y}) = P(\hat{Y}=1|A=a, Y=y) $. EO can be calculated as follow:
\begin{equation}
\label{eq.CV_EO}
\sum_{\forall a^i, a^j \in A}|P_{a^i,0}(\hat{Y})-P_{a^j,0}(\hat{Y})\ | + |P_{a^i,1}(\hat{Y})-P_{a^j,1}(\hat{Y})\ |,
\end{equation}
\label{eq.deo}
then we define $f$ is the probability density function of the confidence score $c$, and thus $f_{a,y}(c) = f(c|A=a, Y=y)$. For the given decision threshold t in range $[0,1]$, we have 
\begin{equation}
\begin{split}
\label{eq.TPR_integral}
    P_{a,y}(\hat{Y}) = \int_{t}^{1} f_{a,y}(c) dc,
\end{split}
\end{equation}
then we can reformulate the Eq. \ref{eq.CV_EO} by the following equation,
\begin{equation}
\resizebox{0.48\textwidth}{!}{$
\begin{gathered}
\label{eq.CV_EO_reformulate}
\sum_{\forall a^i, a^j \in A}\!|P_{a^i,0}(\hat{Y})-P_{a^j,0}(\hat{Y}) | +  |P_{a^i,1}(\hat{Y})-P_{a^j,1}(\hat{Y}))\ |\\
  =\!\sum_{\forall a^i, a^j \in A}|\!\int_{t}^{1}\! f_{a^i,0}(c)-f_{a^j,0}(c) \ dc\ \!|  + |\int_{t}^{1}\!  f_{a^i,1}(c)-f_{a^j,1}(c) \ dc\ \!|.
\end{gathered}
$}
\end{equation}
Therefore, $\forall a^i, a^j \in A$ and $\forall y \in \{0,1\}$, EO is minimized when:
\begin{equation}
\label{eq.pdf_confidence}
f_{a^i,y}(c) = f_{a^j,y}(c).
\end{equation}
Moreover, since the confidence score is the value of logits after applying the sigmoid function, then Eq. \ref{eq.pdf_confidence} is equivalent to $l_{a^i,y}\ \,{\buildrel d \over =}\, \ l_{a^j,y}$, where $l_{a^i,y}$ and $l_{a^j,y}$ are the classifier's distribution of output logits for target class $y$ and sensitive attribute $a^i$ and $a^j$ in the whole logits distribution $l$. Therefore, by minimizing the distribution distance between $l_{a^i,y}$ and $l_{a^j,y}$, we can minimize EO.

\subsection{Motivation}
\label{subsec:motivation}
We see a huge deficiency in the previous logits space regularization methods \emph{Gaussian Assumption} (GA) and \emph{Histogram Approximation} (HA) proposed by \cite{chen2020towards}. Similar EO minimization method discussed in Sec. \ref{subsec:method_fairness_criterion}, \cite{chen2020towards} evaluates the fairness condition with EO and trained a classifier that was invariant to the decision threshold using GA and HA. However, we argue that these two methods were inconsistent with EO minimization, and thus cannot properly achieve fairness. As a result, we leverage the power of MMD, which has a good theoretical property that aligns with the fairness training objectives.


\textbf{Gaussian Assumption} modeled the output logits with Gaussian distributions. That is, the method assumed that optimally $l_{a^i, y} $ and $ l_{a^j, y}$, $\forall y \in \{0, 1\}$, $\forall a^i,a^j \in A$ followed the same Gaussian distributions. Moreover, GA minimized the distance of two Gaussian distributions via symmetric \emph{Kullback–Leibler} divergence, and the detailed implementation can be found in \cite{chen2020towards}. Nevertheless, since there is no clue about the logits distribution, it is not rational to assume the optimal distributions will obey the same Gaussian. As a result, the assumption limits the model's capability of fitting the optimal distribution. 

\textbf{Histogram Approximation} used an N-bins histogram to approximate confidence score distribution. To obtain a differentiable histogram, HA leveraged the Gaussian kernel to approximate the number of samples in each bin and built the distribution of the confidence score by normalizing the item frequency in each bin. Similar to GA, HA minimized the distance of the approximated histogram, i.e., $l_{a^i, y} $ and $ l_{a^j, y}$, $ \ \forall y \in \{0, 1\}$, $\forall a^i,a^j \in A$, by symmetric \emph{Kullback–Leibler} divergence. The detailed setting can be found in \cite{chen2020towards}. Nonetheless, the histogram approximation has the main drawback. \cite{silverman2018density} proved that the optimal bin size depends on the value of density in the bin and the number of sampled data points; therefore, there is no reliable algorithm to construct a histogram, and poor binning will lead to a considerable estimation error in HA.

The above analysis of the drawbacks motivates us to propose a new fairness regularization consistent with EO minimization. Moreover, the experiments in Sec. \ref{sec:results} demonstrate that our new fairness regularization vastly outperforms GA and HA.
\vspace{-10pt}


\section{Method}
\label{sec:method}
\begin{figure*}[ht]
\begin{center}
\includegraphics[width=1.0\linewidth]{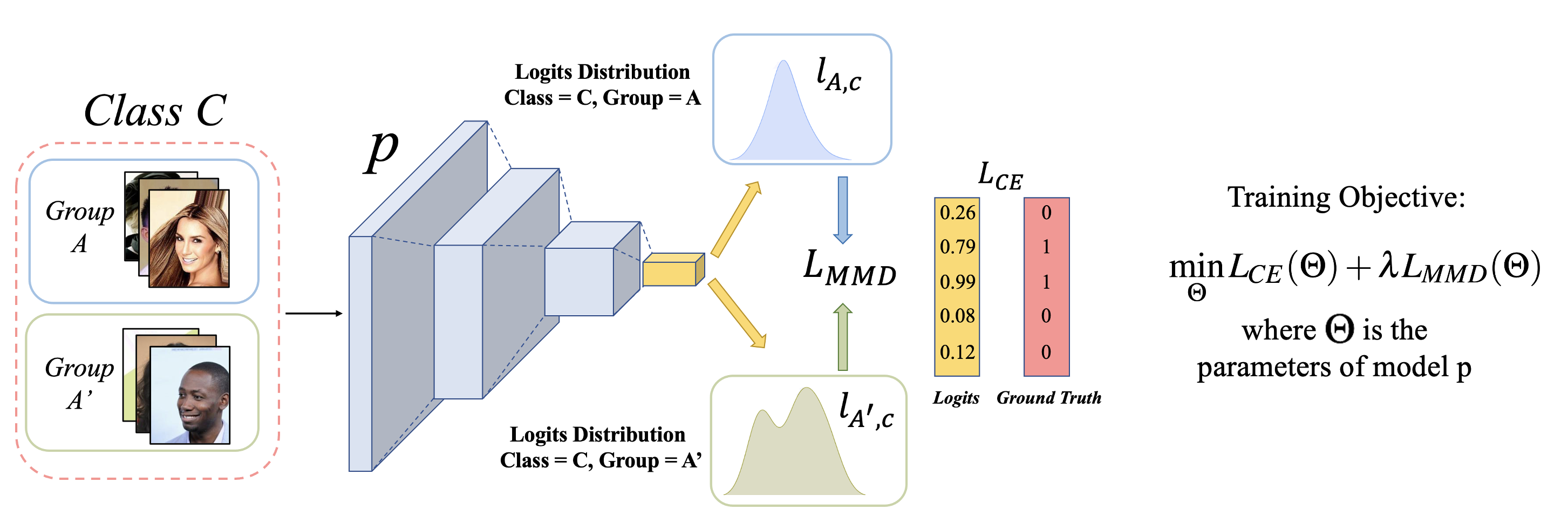}
\caption{The illustrative concept of Logits-MMD. $L_{MMD}$ minimizes the logits distribution discrepancy between each sensitive group for the same class and achieves fairness. $L_{CE}$ is the cross-entropy loss, which enforces the model in achieving high predictive accuracy}
\label{Fig.method}
\end{center}
\end{figure*}

\subsection{Problem Formulation \& Notations}
\label{subsec:method_problem}

In a classification task, define input features $ X = x \in \mathbb{R}^{d}$,
target class $ Y  = y \in\{0,1,2, ..., N \} $, predicted class $\hat{Y}=\hat{y} \in \{0,1,2, ..., N\}$, and sensitive attributes $A  = a \in \{0,1,2, ..., M \} $. In this work, we focus on the case where the target class is binary, i.e., $Y = y \in \{0, 1\}$, $\hat{Y} = \hat{y} \in \{0, 1\}$, and multi-sensitive class, i.e., $A = a \in \{0, 1,2,...,M\}$.
The goal is to learn a classifier $p : x \rightarrow \hat{y}$ that predicts the target class $Y$ to achieve high accuracy while being unbiased to the sensitive attributes $A$ with EO as the fairness metric. 



\subsection{MMD-Based Objective Loss}


To learn the fair classifier in logits space, we leverage Maximum Mean Discrepancy (MMD) \cite{gretton2012kernel}, which measures the largest expected difference between the two sample distributions. Let $\mathbf{D}_{X^{s}} = \{x^s_1, x^s_2, x^s_3, ..., x^s_{n_s}\}$ and $\mathbf{D}_{Y^{t}} = \{y^t_1, y^t_2, y^t_3, ..., y^t_{n_t}\}$ be the sets that sampled from distributions $P(X^s)$ and $Q(Y^t)$. To measures the difference of the given distributions P and Q, MMD is defined as follows:
\begin{equation}
\begin{split}
\label{eq.MMD}
MMD\left (  P, Q\right ) \triangleq &\sup_{f \in \mathcal{H}}\left ( \mathbb{E}_{X^s}[f(X^s)]- \mathbb{E}_{Y^t}[f(Y^t)]\right ).
\end{split}   
\end{equation}   

Through the function $f$ in the reproducing kernel Hilbert space (RKHS) $\mathcal{H}$, MMD is rich enough to distinguish the difference by calculating the mean embedding of the two distributions $P, Q$. The most important theoretical property of MMD that proved by \cite{gretton2012kernel} is
\begin{equation}
\begin{split}
\label{eq.MMD_property}
P=Q \Leftrightarrow MMD ( P, Q) = 0.
\end{split}
\end{equation}

To calculate the distance between $P, Q$ by using MMD defined in Eq. \ref{eq.MMD} efficiently, in practice, we can calculate empirical $MMD\left ( P, Q\right )^{2}$ with the kernel trick, which can be defined as below:

\begin{equation}
\label{eq.MMD_expand}
\begin{split}
    MMD\left ( P, Q\right )^{2}&=\frac{1}{n_{s}^{2}}\sum_{i=1}^{n_{s}}\sum_{j=1}^{n_{s}}k(x^s_{i},x^s_{j})+\frac{1}{n_{t}^{2}}\sum_{i=1}^{n_{t}}\sum_{j=1}^{n_{t}}k(y^t_{i},y^t_{j})\\&-\frac{2}{n_{s}n_{t}}\sum_{i=1}^{n_{s}}\sum_{j=1}^{n_{t}}k(x^s_{i},y^t_{j})
\end{split}
\end{equation}
where $k(\cdot,\cdot)$ is a kernel inducing $\mathcal{H}$. In this work, we use the Gaussian RBF kernel $k(x, x') = \exp (-\frac{1}{2\sigma ^{2}}\left \| x-x' \right \|^{2})$ as our kernel function. 

Based on the EO definition in Sec. \ref{subsec:method_fairness_criterion}, we want to minimize the difference between the logits distribution $ l_{a^i,y}$ and $ l_{a^j,y} $ $\forall y \in \{0,1\}, \forall a^i, a^j \in A$. Therefore, the proposed Logits-MMD regularization $L_{MMD}$ is defined as 
\begin{equation}
\label{eq.logit_MMD_regularization}
\begin{split}
L_{MMD}=\sum_{a^i, a^j \in A}d (l_{a^i,0}, l_{a^j,0}) + d(l_{a^i,1}, l_{a^j,1}).
\end{split}
\end{equation}
$L_{MMD}$ pulls close the pairwise logits distributions  $l_{a^i,y}$ and $l_{a^j,y}$ where
$l_{a,y}$ is the classifier's distribution of output logits for target class $y$ and sensitive attribute $a$ in the whole sampled logits distribution $l$. Moreover, $d$ is the metric to measure the distance of the two sampled distributions. In our case, $d$ is $MMD\left ( P, Q\right )^{2}$. Ultimately EO is minimized when the global optimal of Eq. \ref{eq.logit_MMD_regularization} is achieved i.e. $d (l_{a^i,y}$, $l_{a^j,y}) = 0$, and we get $l_{a^i, y}$ is the same as $l_{a^j, y}$, based on the MMD property described in Eq. \ref{eq.MMD_property}. Therefore, the convergence of our designed fairness objective $L_{MMD}$ is consistent with EO. With the above $L_{MMD}$ regularizer, our training objective is as below:
\begin{equation}
    \label{eq.logit_MMD_objective}
    \min_{\Theta}L_{CE}(\Theta)+\lambda L_{MMD}(\Theta)
\end{equation}
where $\Theta$ is the model parameters and $L_{CE}$ is cross-entropy loss, and $\lambda$ is a tunable hyperparameter that controls the trade-off between accuracy and fairness. The high-level training framework is demonstrated in Fig. \ref{Fig.method}. During the training time, we adopt a stochastic gradient descent optimizer with a mini-batch strategy. For every iteration, we sample out a mini-batch and divide the logits into $|A|$x$|Y|$ sets and optimize the model with the objective in Eq. \ref{eq.logit_MMD_objective}.


\subsection{Comparison with GA \& HA}
\subsubsection{GA}
\begin{figure}[ht]
\begin{center}
\includegraphics[width=0.95\linewidth]{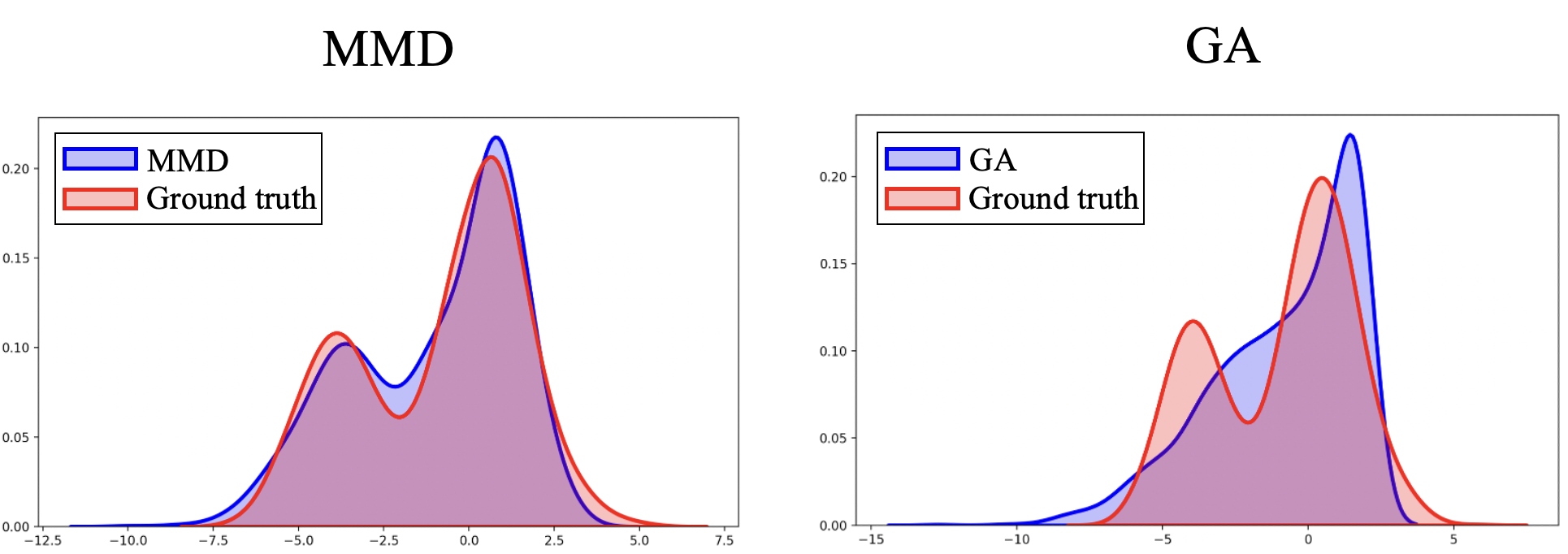}
\caption{The result of GA and MMD fitting a toy example of a multimodal distribution. MMD could fit well on the ground truth toy example, while GA cannot. The x-axis is the value of the data points from the toy example, and the y-axis is the density.}
\vspace{-10pt}
\label{Fig.Toy_example}
\end{center}
\end{figure}

As we introduced in Sec. \ref{subsec:motivation}, the Gaussian Assumption limits the model's capacity to learn complex distribution. However, compared with GA, MMD does not introduce any prior to the distribution of $P, Q$ in Eq. \ref{eq.MMD}; therefore, \textbf{Logits-MMD} has better performance in modeling the target distribution. We provide a toy example to compare GA and MMD. As shown in the Fig. \ref{Fig.Toy_example}. We first construct a simple dataset that targets the model to learn a multimodal distribution. Then, we build a two-layers fully connected network and train it on the toy dataset with GA and MMD only for 500 epochs. It is clear to see that the MMD could fit the target distribution while GA cannot.


\subsubsection{HA}
Compared with HA, MMD avoids density estimating and directly use the statistical mean to measure the distribution discrepancy. Therefore, MMD promises the convergence of the optimization based on the Eq. \ref{eq.pdf_confidence}. On the other hand, the estimation error of kernel density estimation introduces in Sec. \ref{subsec:motivation} hampers the convergence of the HA's optimization.

\section{Experiments}
\label{sec:experiments}

\begin{table*}[ht]
\centering
\caption{Classification results of the fairness (EO) and accuracy (Acc.) evaluation on the test set of the CelebA dataset. We abbreviate the target attribute (T) and the sensitive attributes (S), \emph{Attractive}, \emph{Big Nose}, \emph{Bags Under Eyes}, \emph{Male}, and \emph{Young} as $a$, $b$, $e$, $m$, and $y$, respectively. The results of \emph{LNL}, \emph{MFD}, \emph{FSCL+} are the experimental results reported in \cite{park2022fair}.}
\setlength{\tabcolsep}{9pt}
\label{Table:Compare_SOTA_CelebA}
\footnotesize
\begin{tabular}{c c c c c c c c c c c c c c c}\toprule 
\multicolumn{1}{c}{\text{}} & \multicolumn{2}{c}{\text{T=\emph{a} / S=\emph{m}}} & \multicolumn{2}{c}{\text{T=\emph{a} / S=\emph{y}}} & \multicolumn{2}{c}{\text{T=\emph{b} / S=\emph{m}}} & \multicolumn{2}{c}{\text{T=\emph{b} / S=\emph{y}}} & \multicolumn{2}{c}{\text{T=\emph{e} / S=\emph{m}}} & \multicolumn{2}{c}{\text{T=\emph{e} / S=\emph{y}}} & \multicolumn{2}{c}{\text{T=\emph{a} / S=\emph{m\&y}}}  \\ 
\cmidrule(lr){2-3}
\cmidrule(ll){4-5}
\cmidrule(ll){6-7}
\cmidrule(ll){8-9}
\cmidrule(ll){10-11}
\cmidrule(ll){12-13}
\cmidrule(ll){14-15}
\multicolumn{1}{c}{Methods} & \text{EO} & \text{Acc.} & \text{EO} & \text{Acc.}& \text{EO} & \text{Acc.}& \text{EO} & \text{Acc.}& \text{EO} & \text{Acc.}& \text{EO} & \text{Acc.}& \text{EO} & \text{Acc.} \\ 
\hline
\hline
CNN  & 23.7          & 82.6          & 20.7          & 79.9           & 21.4         & 84.7          & 14.4         & 85.0          & 16.8         & 84.3          & 11.4         & 84.4          & 26.1         & 82.4\\ 
LNL  & 21.8          & 79.9          & 13.7          & 74.3           & 10.7         & 82.3          & 6.8          & 82.3          & 5.0          & 81.6          & 3.3          & 80.3          & 20.7         & 77.7\\ 
HSIC  & 19.4          & 81.7          & 16.5          & 80.3           & 11.2         & 80.8          & 10.5         & 82.6          & 12.5         & 84.0          & 7.4          & 84.2          & 19.3         & 80.1\\ 
MFD  & 7.4           & 78.0          & 14.9          & 80.0           & 7.3          & 78.0          & 5.4          & 78.0          & 8.7          & 79.0          & 5.2          & 78.0          & 19.4         & 76.1\\
FSCL+  & 6.5           & 79.1          & 12.4          & 79.1           & 4.7          & 82.9          & 4.8          & 84.1          & 3.0          & 83.4          & 1.6          & 83.5          & 17.0          & 77.2\\ \hline
TI (HA)  & 16.7          & 81.2          & 17.5          & 79.8           & 14.4         & 81.9          & 10.6         & 85.0          & 10.6         & 84.7          & 6.0          & 85.5          & 21.9         & 82.6\\ 
TI (GA)  & 18.6          & 82.5          & 11.0          & 76.4           & 14.2         & 80.2          & 11.2         & 80.0          & 14.2         & 83.5          & 7.7          & 83.7          & 22.3         & 81.6\\ \hline
Logits-MMD  & \textbf{2.5}  & 80.8          & 11.2          & 79.1           & \textbf{0.5} & 82.4          & \textbf{0.7} & 84.0          & \textbf{2.7} & 83.9          & \textbf{1.3} & 84.5          & \textbf{15.5}         & 81.5 \\ \hline
\end{tabular}
\end{table*}

\subsection{Dataset}

\textbf{\indent CelebA} \cite{liu2018large} 
 The CelebA dataset comprises over 200,000 images, each associated with 40 binary attributes. We follow \cite{park2022fair} and designate \emph{Male} and \emph{Young} as the sensitive attributes. The multi-class sensitive attribute is an element chosen from the set $\{\emph{Male}, \emph{Young}\}$, which forms a two-element subset of the sensitive attribute group. Similarly to \cite{park2022fair}, we consider the target attributes \emph{Attractive}, \emph{Big Nose}, and \emph{Bags Under Eyes}.

\textbf{UTK} \cite{zhang2017age} UTK Face dataset consists of over 20k face images with three annotations, \emph{Ethnicity}, \emph{Age}, and \emph{Gender}. We set \emph{Ethnicity}, \emph{Gender} as the sensitive attribute and the multi-class sensitive attribute is an element chosen from the set $\{\emph{Ethnicity},  \emph{Gender}\}$, which forms a two-element subset of the sensitive attribute group. We set \emph{Age} as the target attribute and redefine \emph{Ethnicity} and \emph{Age} as binary attributes, following \cite{park2022fair}. Since the
UTK dataset doesn’t release official data split for fair evaluation, we split the validation and test set into a balanced dataset for each $\left \{ Sensitive, Target \right \}$ group that has 100 images.



\textbf{Dogs and Cats} \cite{dogNcat}
The Dogs and Cats dataset has 38500 dog and cat images. Besides the original species labels (dogs or cats), LNL\cite{kim2019learning} annotated the color labels (bright or dark). We set the \emph{color} to the sensitive attribute and the \emph{species} to the target attribute. We organize completely balanced validation and test sets for each $\left \{ Sensitive, Target \right \}$ group with 600 images. Besides, We compose a color-biased dataset, where a sensitive group(\emph{e.g} Black) has cat data $N$ times as much dog data. In contrast, the other sensitive group has the opposite color ratio. The $N$ is set to 6, 5, 4, and 3 to simulate different bias levels. 

\subsection{Implementation Details and Evaluation Protocol}
We utilize ResNet-18 as our baseline CNN. During the training phase, the data is augmented by random flipping, rotation, and scaling. The network is trained for 200 epochs using SGD optimizer with an initial learning rate set as 0.01. Hyperparameter $\lambda$ for the regularization term is set 
in the range from 0.01 to 0.1 in each experiment. We followed the original paper or released code for the other baseline methods and reproduced them in Pytorch. To evaluate the accuracy and fairness of our proposed method, we used top-1 accuracy (Acc.) and equalized odds (EO) metrics on a different dataset. Specifically, we followed the approach of \emph{FSCL+} \cite{park2022fair} for calculating these metrics.

\section{Results}
\label{sec:results}
\subsection{Baseline}
In this section, we compare Logits-MMD with several state-of-the-art: LNL \cite{kim2019learning}, HSIC \cite{quadrianto2019discovering}, MFD \cite{jung2021fair}, and FSCL+ \cite{park2022fair}. CNN is our baseline, where the objective is to learn an accurate classifier with cross entropy loss only. We also compare the results with similar logits space regularization methods, ``Gaussian Assumption (GA)” and ``Histogram Approximation (HA)” proposed by \cite{chen2020towards}. The experimental results for the CelebA and UTK Face datasets are demonstrated in Table \ref{Table:Compare_SOTA_CelebA} and Table \ref{Table:Compare_SOTA_UTK}, respectively. 


\subsection{Experimental Results on CelebA}
For the CelebA dataset, in Table \ref{Table:Compare_SOTA_CelebA}, we follow the sensitive and target groups setting in \cite{park2022fair}. The CNN baseline has relatively high accuracy but poor performance in fairness. We focus on comparing our proposed Logits-MMD with FSCL+, which has the best trade-off between top-1 accuracy and EO in all state-of-the-art methods. Compared with FSCL+, our method improves EO by 40.6\% and the accuracy by 3.9\% on average. Moreover, our method significantly outperforms GA and HA. This phenomenon indicates that the MMD objective is more consistent with the EO the two methods.

\begin{table}[ht]
    \centering
    \caption{Classification results of the fairness (EO) and accuracy (Acc.) evaluation on the UTK dataset. The target attribute (T) $A$ stands for Age, and the sensitive attribute (S) $E$, $G$ stand for Ethnicity and Gender. }
    \setlength{\tabcolsep}{2pt}
    \label{Table:Compare_SOTA_UTK}
    \begin{tabular}{c||c|c||c|c}
    \hline
    {Methods}       & \multicolumn{2}{c||}{T=A / S=E}   & \multicolumn{2}{c}{T=A / S=G}  \\ \cline{2-5}
                                   & EO            & Acc.              & EO           & Acc.          \\ \hline \hline
    CNN                           & 11.6          & 90.7              & 13.5         & 90.5          \\ 
    LNL                            & 6.3           & 90.3              & 8.3          & 90.0          \\ 
    HSIC                          & 10.0          & 90.5              & 9.6          & 90.3          \\ 
    MFD                            & 9.4           & 90.0              & 10.9         & 90.3          \\
    FSCL+                          & 5.6           & 90.3              & 3.5          & 90.5          \\ \hline
    TI (HA)                        & 8.8           & 91.3              & 8.5          & 89.8          \\ 
    TI (GA)                        & 9.9           & 90.5              & 10.9         & 90.5          \\ \hline
    Logits-MMD                      & \textbf{4.8}  & 90.0              & \textbf{3.2} & 88.5          \\ \hline
\end{tabular}
\end{table}
\begin{figure*}[ht]

\begin{center}
\includegraphics[width=0.99\linewidth]{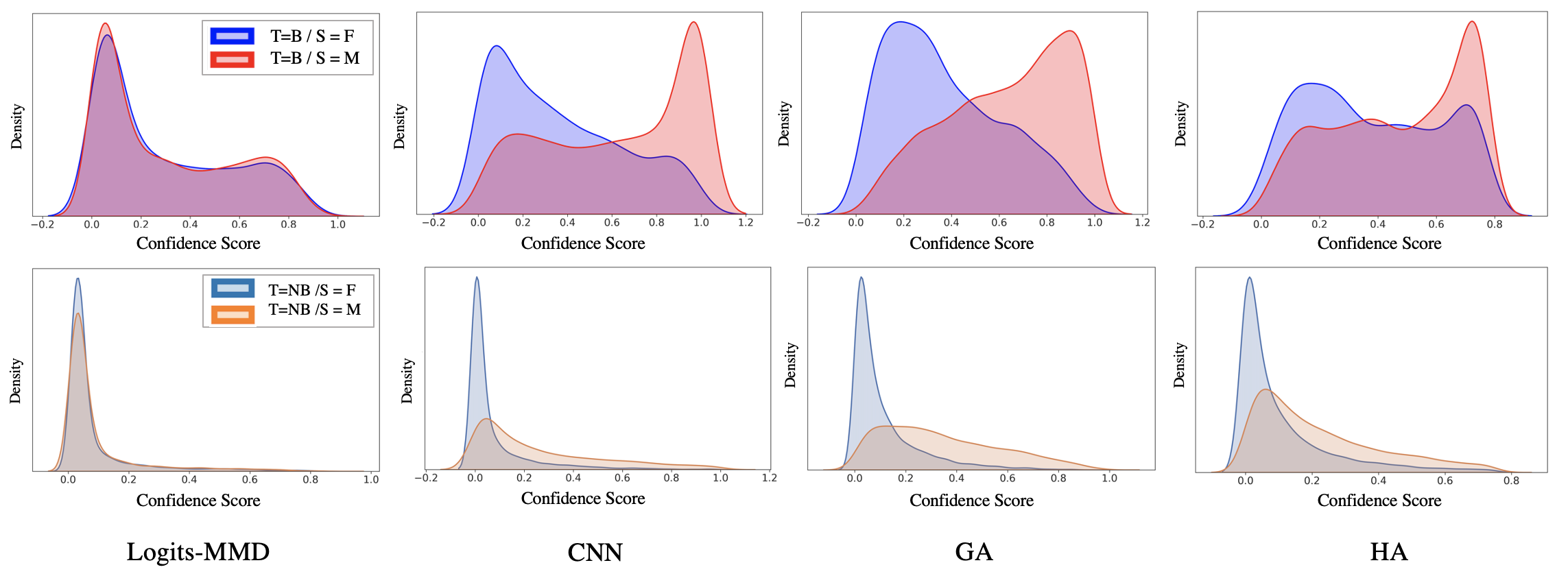}
\caption{ The first row is the confidence score PDF with the ground truth as $Big \ Nose $ (B), and the second is the ground truth as $Not \ Big \ Nose$ (NB). From left to right are Logits-MMD, CNN, Gaussian Assumption (GA), and Histogram Approximation (HA), respectively.``T'' refers to the target attribute, and ``S'' to the sensitive attribute. The density estimation will smooth the curve at the max and min points, so there will be values that exceed 1 and less than 0.}
\vspace{-10pt}

\label{Fig.visualize_tmp}
\end{center}
\end{figure*}

\subsection{Experimental Results on UTK}
For the UTK Face dataset, we reproduce the results of all the other state-of-the-art models and compare our results with them. For different sensitive attribute \emph{Ethnicity}, \emph{Gender} and $\{\emph{Ethnicity}, \emph{Gender}\}$, in Table \ref{Table:Compare_SOTA_UTK}, our method has the improvement in EO by 14.3\%, 8.5\% and 16.4\%, respectively, compared with FSCL+. Similar to the results of CelebA dataset, Logits-MMD significantly outperforms GA and HA in terms of fairness while accuracy drops slightly.

\begin{table}[ht]
    \centering
    \caption{Classification results of the fairness (EO) and accuracy (Acc.) evaluation on our own split test set. $N$ refer to different bias level in the train set. In this experiment, target attribute is species and sensitive attribute is color.}
    \setlength{\tabcolsep}{1pt}
    \label{Table:Compare_SOTA_dogNcat}
    \begin{tabular}{c||c|c||c|c||c|c||c|c}
    \hline
    {Methods}  & \multicolumn{2}{c||}{$N$ = 6} & \multicolumn{2}{c||}{$N$ = 5}& \multicolumn{2}{c||}{$N$ = 4} & \multicolumn{2}{c}{$N$ = 3}  \\ \cline{2-9}
                      & EO            & Acc.          & EO                      & Acc.          & EO           & Acc.          & EO           & Acc.          \\ \hline \hline
    CNN                   & 3.67          & 97.67          & 3.42                   & 97.79          & 2.58         & 98.13          & 1.58         & 98.38          \\ 
    LNL                   & 3.42          & 97.46          & 3.17                    & 97.58         & 2.25          & 97.88          & 1.58         & 98.04          \\ 
    HSIC                 & 3.17          & 97.50          & 3.33                   & 97.58          & 2.33         & 97.92          & 1.25          & 98.13          \\ 
    MFD                   & 3.25           & 96.54          & 2.58                    & 97.95          & 2.25          & 97.71          & 0.92          & 97.75          \\
    FSCL+                 & 3.00           & 95.42          & 2.92                    & 95.46          & 2.17          & 95.67          & 0.75          & 96.21          \\ \hline
    TI (HA)                & 3.17          & 97.41          & 2.50                   & 97.67          & 2.33         & 97.66          & 1.33          & 98.25          \\ 
    TI (GA)                & 3.25          & 97.29          & 3.00                   & 97.75          & 2.42         & 97.88          & 1.42          & 98.29         \\ \hline
    Logits-MMD             & \textbf{1.50}  & 97.00          & \textbf{1.33}          & 97.75         & \textbf{0.83} & 98.08          & \textbf{0.62} & 98.33         \\ \hline
\end{tabular}
\end{table}

\subsection{Experimental Results on Dogs and Cats}
To check that our method is generalizable to mitigate various types of bias, we set different bias levels based on the color setting from \cite{kim2019learning} in the Dogs and Cats dataset by the hyperparameter $N$. We also reproduce the results of all the other state-of-the-art models and compare our results with them. In Table \ref{Table:Compare_SOTA_dogNcat}, $N$ refers to black cat data having $N$ times as much dog data while the white animal group has an opposite setting. Logits-MMD has the best performance regarding the trade-off between accuracy and fairness when the dataset suffers from different degrees of bias. This result demonstrates our method's feasibility in different scenarios of bias setting.

\subsection{Qualitative Analysis of Logits-MMD}
In this subsection, we compare the qualitative result of Logits-MMD with the similar logits space regularization method GA, HA, and CNN. We visualize the probability density function (PDF) of CelebA dataset under the setting $T=b$ and $S=m$ since it has the most imbalanced data setting. In this setting, 10949 images belong to group $\left \{ Not\ Big\ Nose, Female \right \}$, 4781 to $\left \{ Not\ Big\ Nose, Male \right \}$, 1298 to $\left \{ Big\ Nose, Female \right \}$, and 2934 to $\left \{ Big\ Nose, Male \right \}$; moreover, their corresponding PDF of confidence score are $f_{nb,f}(c)$, $f_{nb,m}(c)$, $f_{b,f}(c)$, and $f_{b,m}(c)$ respectively.  The first row of Fig. \ref{Fig.visualize_tmp} is the visualization of $f_{b,f}(c)$ and $f_{b,m}(c)$, and the second row is $f_{nb,f}(c)$ and $f_{nb,m}(c)$.  For GA, HA, and CNN, it is clear to see that the model tends to predict the $ Female$ images to be $Not \ Big \ Nose$ and the $Male$ ones to be $Big \ Nose$ as there is a peak on the left for $f_{b,f}(c)$ and a peak on the right for $f_{b,m}(c)$. This phenomenon is caused by the imbalanced dataset setting. Nevertheless, Logits-MMD has the most similar logits distribution between the different sensitive groups' visualization results compared with GA, HA, and CNN. This result demonstrates the Logits-MMD's capability to achieve fairness under the EO criterion. 




\section{Conclusion}
\label{sec:conclusion}
In this paper, we proposed a novel training framework in which the fairness regularization constraint was consistent with the fairness criterion. With our theoretical proof and extensive experiments, we demonstrated that our method outperformed similar logits space regularization and achieved state-of-the-art on two facial attribute datasets. Moreover, we showed that our proposed method had good potential to mitigate different degrees of general bias through the experiment on the Dogs and Cats dataset.

\bibliographystyle{ACM-Reference-Format}
\bibliography{sample-sigplan}










\end{document}